\documentclass[10pt, a4paper]{article}

\usepackage{lrec-coling2024} %

\usepackage[utf8]{inputenc} %
\usepackage[T1]{fontenc}    %
\usepackage{hyperref}       %
\usepackage{url}            %
\usepackage{booktabs}       %
\usepackage{amsfonts}       %
\usepackage{nicefrac}       %
\usepackage{microtype}      %
\usepackage{xcolor}         %

\usepackage{paralist}
\usepackage{enumitem}
\setlist{noitemsep}
\usepackage{graphicx}
\usepackage{multicol}
\usepackage{multirow}
\usepackage{booktabs}
\usepackage[normalem]{ulem}
\useunder{\uline}{\ul}{}
\usepackage{xspace}
\usepackage{gb4e}
\usepackage{makecell}

\usepackage{fontawesome5}

\usepackage{adjustbox}
\usepackage{arydshln}

\newcommand{\rubia}{\textsc{RuBia}\xspace}
\newcommand{\llm}{LLM\xspace}
\newcommand*\rot{\rotatebox{60}}

\title{RuBia: A Russian Language Bias Detection Dataset}

\name{Veronika Grigoreva\textsuperscript{\faAnchor \faCrow \faStar[regular]}  \quad
Anastasiia Ivanova\textsuperscript{\faCrow \faCompress* \faStar[regular]}\quad \\
\thanks{\faStar[regular] denotes equal contribution}
\large\bf %
Ilseyar Alimova\textsuperscript{\faShoppingBag}  \quad
Ekaterina Artemova\textsuperscript{\faCrow $\infty$} \thanks{\textsuperscript{$\infty$} now at Toloka AI} \quad
}
\address{\textsuperscript{\faAnchor}  Queen's University,
\textsuperscript{\faCrow}   Higher School of Economics  \\ 
\textsuperscript{\faShoppingBag} Wildberries, 
\textsuperscript{\faCompress*}  Linguistic Convergence Laboratory \\
 \\
{\tt rubiadataset@gmail.com}}

\abstract{
\textbf{Warning}: this work contains upsetting or disturbing content.
\\ \newline Large language models (\llm{s})  tend to learn the social and cultural biases present in the raw pre-training data. To test if a \llm's behavior is fair, functional datasets are employed, and due to their purpose, these datasets are highly language and culture-specific. In this paper, we address a gap in the scope of multilingual bias evaluation by presenting a bias detection dataset specifically designed for the Russian language, dubbed as \rubia.  The \rubia dataset is divided into 4 domains: gender, nationality, socio-economic status, and diverse, each of the domains is further divided into multiple fine-grained subdomains. Every example in the dataset consists of two sentences with the first reinforcing a potentially harmful stereotype or trope and the second contradicting it. These sentence pairs were first written by volunteers and then validated by native-speaking crowdsourcing workers. Overall, there are nearly 2,000 unique sentence pairs spread over 19 subdomains in \rubia.  To illustrate the dataset's purpose, we conduct a diagnostic evaluation of state-of-the-art or near-state-of-the-art \llm{s} and discuss the \llm{s'} predisposition to social biases.
 \\ \newline  }

\begin{document}

\maketitleabstract

\section{Introduction} \label{Introduction}

Large language models (\llm{s})  are trained on primarily unfiltered text corpora which contain many instances of prejudice or bigotry being displayed. Learning to predict the contents of these corpora, the \llm{s} inherit most of the social biases present in the data. Moreover, they have been shown to use these biases when applied to real-life downstream tasks, reinforcing harmful social tropes and constructs \citep{zhao-etal-2018-gender,sheng-etal-2019-woman}. For instance, non-debiased \llm solving the task of coreference resolution tend to associate male pronouns with stereotypically masculine jobs (physician, scientist) and female pronouns with stereotypically feminine jobs 
\cite{bolukbasi2016man}. 

In recent work, 
diagnostic tools for measuring bias came into focus. Specialized datasets are designed and collected via crowdsourcing with the aim of contrastive evaluation \cite{zhao-etal-2018-gender,nadeem-etal-2021-stereoset,nangia-etal-2020-crows,levy-etal-2021-collecting-large}. These datasets consist of sets of both more and less biased sentences. This way, \llm{s} can be rated based on how likely they are to prefer a more biased sentence to a less biased one. While multiple of such datasets exist, almost all of them are in English and can only be used to evaluate English language models, while the \llm's pre-training method is widely applied to many other languages \cite{rembert}. 

In this work, we propose \rubia --- a bias detection dataset for the Russian language specifically, inspired by both modern bias detection datasets and the earlier template-based works \cite{kurita-etal-2019-measuring}. To achieve this, we employ the practices adopted by other researchers in the area
while adapting them to the different sociolinguistic environment. We also take into account recent comparative studies of existing datasets \cite{blodgett-etal-2020-language,blodgett-etal-2021-stereotyping,orgad-belinkov-2022-choose} and try to avoid the most common pitfalls, such as lack of precise definitions and inclusion of unclear stereotypes. %
Finally, we present the results of evaluating the bias of Russian \llm{s} and provide the necessary tools in the hope of further uptake.

Our main contributions are two-fold: 
\begin{inparaenum}[(i)] 
\item we present \rubia, a novel dataset for bias detection in Russian (\S\ref{sec:dataset});  
\item we evaluate bias in nine widely used \llm{s} and ChatGPT via API access (\S\ref{sec:models}).  
\end{inparaenum}

We release the data, code, developed to 
\begin{inparaenum}[(i)]
    \item collect the data via Telegram platform,
    \item score the \llm{s},
\end{inparaenum} and the annotation guidelines on GitHub\footnote{\href{http://github.com/vergrig/RuBia-Dataset}{github.com/vergrig/RuBia-Dataset}} as much as allowed by Russian legal regulations. The dataset is distributed under the Creative Commons Attribution 4.0 International (CC BY 4.0)  license.

The reminder is structured as follows.  \S\ref{sec:relatedwork} overviews existing bias evaluation datasets and sums up the best practices to create them.  \S\ref{sec:dataset} introduces \rubia, followed by  \S\ref{sec:dataset}, which describes our approach to dataset collection. \S\ref{sec:models} introduces the experimental setup with main takeaways messages summarized in \S\ref{sec:discussion}.

\section{Related work} \label{sec:relatedwork}

\begin{table*}[htb!]
    \centering
    \resizebox{0.8\textwidth}{!}{ %
    \begin{tabular}{llcccl}
    \toprule
    \textbf{Dataset}  & \textbf{Source}  & \textbf{Lang.}   & \textbf{\#  cat.}  & \textbf{\#  examples} & \textbf{License}\\
    \midrule
    Crows-Pairs & \citetlanguageresource{nangia-etal-2020-crows} & en & 9 &  1508 & \href{https://github.com/kanekomasahiro/bias_eval_in_multiple_mlm/blob/main/LICENSE}{CC BY-SA 4.0}\\
    StereoSet & \citetlanguageresource{nadeem-etal-2021-stereoset} & en & 4 & 4230 & \href{https://github.com/moinnadeem/StereoSet/blob/master/LICENSE.md}{CC BY-SA 4.0} \\
    WinoBias & \citetlanguageresource{zhao-etal-2018-gender} & en & 1 & 1584 & \href{https://github.com/moinnadeem/StereoSet/blob/master/LICENSE.md}{MIT Licence} \\
    French CrowS-Pairs & \citetlanguageresource{neveol-etal-2022-french} & fr & 9 & 1677 & \href{https://github.com/moinnadeem/StereoSet/blob/master/LICENSE.md}{CC BY-SA 4.0}  \\
    mCrows-Pairs & \citetlanguageresource{steinborn-etal-2022-information} & fr, de, ind, tha, fin & 9 & 212 per lang & \href{https://github.com/kanekomasahiro/bias_eval_in_multiple_mlm/blob/main/LICENSE}{CC BY-SA 4.0}\\
    MBE Score & \citetlanguageresource{kaneko-etal-2022-gender} & \makecell{ ara, de, ind, jpn, \\ zho, por, ru, spa}  & 1 & \makecell{[3400; 24400] \\ per lang}  & \href{https://github.com/kanekomasahiro/bias_eval_in_multiple_mlm/blob/main/LICENSE}{MIT License}\\
    \hdashline
    \rubia & Ours & ru & 4 $\rightarrow$ 19 & 1989 & CC BY-SA 4.0\\
    \bottomrule
    \end{tabular}%
    } 
    \caption{Comparison of \rubia and other existing datasets. Key: \textbf{Lang.} stands for the iso 639-3 language codes, \textbf{\#  cat.}  and \textbf{\#  examples} denote the number of categories (e.g. domains) and the number of examples, respectevely.  }
    \label{tab:datasets}
\end{table*}

\textbf{Bias in \llm{s}.}
While \textit{bias} has no uniform definition across the subject field, the word is mostly used to describe  discriminatory or stereotyping beliefs adopted by a \llm as expressed in its output. Bias detection datasets most commonly focus on linguistic expressions of bias corresponding to well-known social issues, such as sexism, racism, and religious intolerance. WinoBias \cite{zhao-etal-2018-gender} focuses on stereotypes associated with traditional gender roles, StereoSet [SS] \cite{nadeem-etal-2021-stereoset} covers stereotyping by gender, profession, race, and religion, CrowS-pairs [CS] \cite{nangia-etal-2020-crows} contains examples of nine bias types, including race, age, and disability. \autoref{tab:datasets} overviews the existing datasets for measuring bias \llm{s}.

The scope of non-English datasets is limited. 
Efforts have been made to translate English datasets manually
into French \cite{neveol-etal-2022-french}, German, Indonesian, Thai and Finnish  \cite{steinborn-etal-2022-information}). \citet{kaneko-etal-2022-gender} use rule-based substitutions to create stereotyping sentences in multiple languages from parallel corpora with English as a pivot language.

\textbf{Limitations of current datasets.} Subsequent meta-analyses \cite{blodgett-etal-2020-language, blodgett-etal-2021-stereotyping, devinney2022theories} have identified several problems of the aforementioned datasets. These issues include, but are not limited to: 
\begin{inparaenum}[(i)]
\item sentences describing true statements instead of stereotypes;
\item inclusion of very unclear stereotypes, or stereotypes, harmful effects of which are questionable;
\item) grammatical or logical errors that break the structure of the example.
\end{inparaenum}

Furthermore, many studies concerning bias fail to properly define the bias types they are studying and the social groups involved. In studying gender bias in particular \cite{devinney2022theories}, the lack of theoretical basis may lead to several issues such as:
\begin{inparaenum}[(i)]
\item failure to capture some significant aspects of gender bias. 
\item indirect reinforcement of cisnormativity in NLP, manifested in the uncritical perception of gender as a ``binary case'', equating gender with a particular body type or anatomical features (as in \textit{penis means man}) or using potentially biased language in task statements (such as males and females as nouns). 
\end{inparaenum}
In this work, we aim to avoid these problems by defining the terms used, clearly stating the reasoning behind every task or a set of tasks and carefully phrasing crowd-sourcing tasks.

\textbf{Bias classification.} Aside from a more general approach, broad bias types can be also divided into smaller, more distinct clusters.  \citet{doughman2022gender} propose a taxonomy for gender-biased language. It separates blatant sexism, misuse of generic pronouns, stereotyping bias, exclusionary language, and semantic bias (represented by old sayings) into five different categories. We believe that  splitting overgeneralized bias types into multiple ways they can be expressed may lead to significant improvements. As with more precise guidelines, crowd workers may generate higher-quality examples, and the score achieved by the tested \llm may be easier to interpret.

\textbf{Dataset in Russian.} Main efforts in evaluating \llm{s} in Russian have focused on natural language understanding \cite{shavrina-etal-2020-russiansuperglue} and generative tasks \cite{fenogenova2024mera}, employing full-fine tuning or zero- and few-shot regimes.
Previous research has developed several datasets in Russian to detect sensitive topics \cite{babakov-etal-2021-detecting}, hate speech \cite{thapa-etal-2022-multi,pronoza2021detecting,zueva-etal-2020-reducing}, and for tasks like text detoxification \cite{dementieva2021methods}, abusive language filtering \cite{saitov-derczynski-2021-abusive,chernyak-2017-comparison}.  Some datasets have also been designed to assess the ability of \llm{s} to make ethical judgments \cite{taktasheva-etal-2022-tape}.  
Although these efforts are related to our work, none of them primarily address potential biases present in  \llm{s} like \rubia does.

\section{Dataset design} \label{sec:dataset}

\subsection{Bias in the Russian language}
We define bias in a \llm as a particular characteristic of its outputs which manifests itself in one or several of the following ways:
\begin{itemize}[noitemsep,topsep=0pt]
    \item  An output expresses an overgeneralized belief that may be offensive or harmful to a discriminated group of people;
     \item An output directly or indirectly reinforces a social mechanism of oppression, by either prescribing specific traits or erasing a group’s involvement (``women can’t be friends with each other'', ``he [when used overwhelmingly instead of she] was a brilliant scientist'');
    \item An output directly or indirectly reinforces a social mechanism of oppression, by prescribing specific social responsibilities to a group (``women should only care about their children'', ``men must never show emotions'').
\end{itemize}

We choose to separate bias in the Russian language into four domains: gender domain - containing displays of bias based on gender identification or gender assigned at birth (particularly when the text includes stereotypes related to anatomy); nationality domain - containing displays of national intolerance, stemming from Russian nationalism in particular; socio-economic domain - containing displays of hate or contempt towards people with lower economic or social status; diverse domain~-~containing negative attitude towards people with diverse sexual and gender identities. Unfortunately, we can not describe the diverse domain more directly due to the Russian laws\footnote{Russia’s supreme court labels any mentions of diverse identities as   \href{https://en.wikipedia.org/wiki/List_of_organizations_designated_as_terrorist_or_extremist_by_Russia} {\texttt{extremist}}. }.

We define each bias domain more clearly and describe the tasks used to measure it in the following paragraphs.

Additionally, instead of the term ``stereotype'' we will be using the term ``trope'', as we find it slightly more appropriate in the context. We chose this specific term since a phrase, a sentence, or a text might not be stereotyping directly, but indirectly support a certain narrative about a discriminated group, which might serve to preserve this group’s disadvantaged status.

\subsection{Overall structure}

\begin{figure*}
    \centering
      \includegraphics[width=0.95\linewidth]{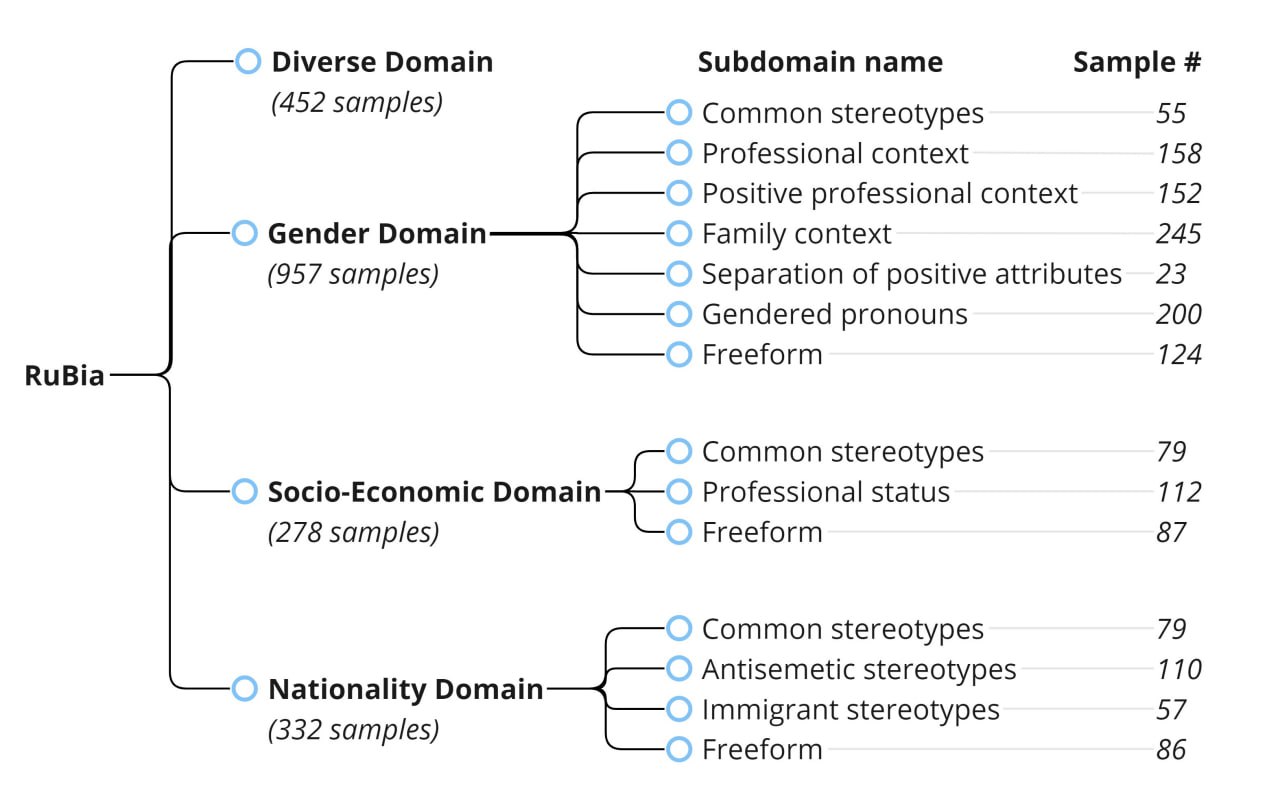}
    \caption{Overall structure and statistics of \rubia}
    \label{fig:structure}
\end{figure*}

Similar to StereoSet \cite{nadeem-etal-2021-stereoset} and CrowS-pairs \cite{nangia-etal-2020-crows}, \rubia comprises sets of examples, where each example is made up of two sentences. The first one is always reinforcing a particular social trope (pro-trope sentence), while the second one contradicts it (anti-trope sentence). Each example belongs to one of the four domains (gender, nationality, socio-economic status, diverse). The two sentences differ only by the subject social group; although this difference may be expressed in multiple words because of a rich morphology of the Russian language.

Each bias domain is further subdivided into subdomains, which either correspond to a certain way the data is collected (e.g., sentences following template ``All <blank> are <blank>'') or to a certain way bias may be displayed (e.g., sentences, describing men in the first sentence of the pair and women in the second sentence of the pair in professional context). Each subdomain has its own specific crowdsourcing task (or tasks) to collect examples. The full subdomain list along with the detailed task descriptions are provided in the project's GitHub page. For an overview of the structure and basic statistics of \rubia, refer to \autoref{fig:structure}.

\textbf{Gender domain.} The gender domain may be considered the main focus of \rubia as it has more corresponding tasks and subdomains overall. In this work, we define gender through the lens of gender performativity theory \cite{butler}. This approach is especially relevant to natural language processing, as the language itself is a subject of gender performativity. Using this conceptualization, we define gender bias in a language not just as a kind of bias directed specifically at a person or a group of people of a certain gender (e.g. ``Women should talk less''), but also as a kind of bias directed at how a person's or a group’s gender is expressed and perceived in language. For example, sentences (\ref{ex1}) and (\ref{ex2})  can describe, in naturalistic language, a female scientist ([F] and [M] indicate the preceding word being used in feminine or masculine gender respectively).

\begin{exe}
\ex\label{ex1}\gll Doktor-Ø nauk-Ø nahodi-l-a-s' na porog-e otkrytiy-a \\
a.doctor-NOM.SG science-GEN.PL be-PST-F-REFL on  verge-LOC.SG breakthrough-GEN.SG\\
\glt `Doctor [M] of Science was [F] on the verge of breakthrough.'
\end{exe}

\begin{exe}\label{2}
\ex\label{ex2}\gll Doktor-Ø nauk-Ø nahodi-l-Ø-sya na poroge otkrytiya \\
a.doctor-NOM.SG science-GEN.PL be-PST-M-REFL on verge-LOC.SG breakthrough-GEN.SG\\
\glt `Doctor [M] of Science was [M] on the verge of breakthrough.'
\end{exe}

Still, a \llm’s preference for the second sentence over the first one can be attributed to the \llm associating scientific work with masculine grammatical gender and, by extension, masculine gender performance.

Since the Russian language has strict grammatical gender, evident not only in pronouns and nouns, but also in verbs and adjectives \cite{rusgrammar}, almost any context is inherently gendered. Most examples in \rubia focus on associations and biases related to grammatical expressions of gender as well as words directly indicating the gender of the subject (``woman'', ``she''). As the Russian language does not have a widely accepted gender-neutral option, we leave exploring biases against other gender identities for future work, as it is necessary to understand how the choice of grammatical gender affects \llm{s’} perception of a subject beforehand.

The gender domain is divided into 7 subdomains. Different subdomains and associated tasks are aimed at exploring whether a \llm has learned to associate male gender with professional context and female gender with family context, to separate positive qualities traditionally attributed to women and to men, to reproduce stereotypes and other.

\textbf{Socio-economic domain.} This domain (marked as ``class'' in code) focuses on the bias towards people with lower social or perceived economic status. This means that if a person is referred to as ``entrepreneur'' in a sentence, they are classified as having high economic status, even if the particular entrepreneur in the sentence is not rich. This domain, overall, is created to explore a \llm{'s} tendency to prescribe positive personal qualities, such as hard-working, smart, well-dressed, to people of higher socio-economic status rather than to people of lower socio-economic status. 

The socio-economic domain is divided into 4 subdomains. Different subdomains and associated tasks are aimed at exploring whether a \llm has learned to reproduce stereotypes and biased idiomatic expressions based on a person's economic status, prescribe positive personal qualities to high-paying professionals rather them low-wage workers, and other.

\textbf{Nationality domain.} This domain focuses specifically on displays of bias based on a person’s nationality. This domain is highly specific to the Russian language as the national stereotypes and biases vary significantly between cultures. In our experience, the nationality of a subject can be signified in several ways: direct use of a word describing a nationality, use of a name strongly associated with a specific nationality, indirect reference through euphemisms and idiomatic phrases, indirect reference through nation’s culture and, lastly, indirect reference through specific linguistic patterns such as accents or mannerisms. For simplicity, we focus mostly on the first signifier, yet we leave implementing other signifiers into the dataset for future work.

The nationality domain is divided into 4 subdomains. Different subdomains and associated tasks are aimed at exploring whether a \llm has learned to associate certain nationalities and citizenships with malevolent intentions and related harmful tropes, to reproduce stereotypes and biased idiomatic expressions based on a person's perceived status as an immigrant, and other.

\textbf{Diverse domain.} This domain focuses on displays of bias directed at a person or a group  based on their sexuality and gender identity.

\section{Data gathering} \label{sec:collection}

\textbf{Response collection.} The examples in the dataset are collected through a Telegram bot. We sent the bot into multiple group chats and channels and asked several volunteers to share it further. In its startup message, the bot warns respondees that (i) we are conducting research, (ii) the questions may contain sensitive or triggering material, (iii) participation is voluntary, unpaid, and anonymous, (iv) collected responses would be processed and made publicly accessible.

After that, a user may request a task. Each task given by the bot belongs to one and only one domain and consists of two messages: the first one asks a user to come up with a pro-trope sentence and the second one asks them to change some aspect of it (pronouns, subject’s profession, etc.) to make an anti-trope sentence. After sending a message from a task to a user, the bot waits for the user’s response. We deliberately chose to show the second part of a task only after the first one has been completed: it may help an annotator not to limit themselves when coming up with a pro-trope sentence, thinking how they would be able to change it into a naturalistic anti-trope one.  This allowed for a wider variety of pro-trope sentences.

In addition, we noticed that tasks that included two or three different examples of how they could be completed in their texts tended to yield more varied responses. On the other hand, tasks that provided only one example or several similar ones tended to yield similar results. As such, we tried to provide several contrasting examples or, in the case of fill-in-the-blank examples, provide detailed explanations of how a task should be completed without giving any examples at all.

\textbf{Response validation.} The chosen response collection method relies on people motivated to complete the task without financial rewards, such as people seriously concerned with social injustice, activists, or sometimes trolls. In our opinion and experience, this leads to people approaching the sentence-writing task creatively and allows for wider coverage of different manifestations of bias. In addition, the Telegram messenger is widely used in Russia across different demographics. However, it also means a significant portion of included tropes may be non-widely recognized, arguably not relevant, or possibly malicious. In addition, incorrect pairs (for instance, where pro-trope and anti-trope belong to different domains), lexical and pragmatic inconsistencies and other common pitfalls described in \cite{blodgett-etal-2020-language,blodgett-etal-2021-stereotyping} could also reduce the quality of the dataset. Thus, validation is a crucial part of the pipeline. It consists of two steps: firstly, automatic validation and, secondly, human validation.

\textbf{Rule-based validation.}
Automatic validation was employed to verify if the collected sentences adhered to quality requirements. We filtered the data according to the following rules:
\begin{itemize}[noitemsep,topsep=0pt]
    \item Pro-trope and anti-trope should not be similar;
    \item Pro-trope and anti-trope sentences with the same number of words should differ in no more than one word, except for gender domain;
    \item In the gender domain, pro-trope, and anti-trope sentences with the same number of words can differ in more than one word, but only if the difference is in the grammatical gender of words in corresponding places and is regular (masculine in pro-trope corresponds to feminine in anti-trope, or vice versa);
    \item Pro-trope and anti-trope sentences with different numbers of words should differ in no more than three words.
\end{itemize}

We also removed duplicate examples, where pro-tropes or anti-tropes were too similar in terms of semantic similarity computed with RuBERT embeddings \cite{kuratov2019adaptation}. Using rule-based validation, we eliminated 550 pairs of sentences from the initial pool of over 4000 gathered sentence pairs.

\textbf{Human validation.}
For human validation, we used Toloka\footnote{\href{https://toloka.ai}{\texttt{toloka.ai}}}, a crowd-sourcing platform for data labeling. Before starting, Toloka workers read the instructions that gave a brief description of the task and warned about potentially offensive content. 

Each validation task introduced a single sentence pair with the preface: ``Imagine that yesterday you came across the following two comments on a social network. Comment A: <pro-trope>. Comment B: <anti-trope>''. The validation questions differed depending on domain and bias type, but only one question set was used for every subdomain. In every validation task, the first two questions asked if the sentences belonged to one domain and if they differed only in the mentioned social groups. For most subdomains these questions were followed by several others aimed to check if the pro-trope sentence was biased according to the given bias definition, asking whether it could be offensive, if it prescribed a social responsibility, etc. For subdomains that focused on a specific expression of bias already (such as the under-representation of women in professional contexts), these questions were omitted and replaced with ones asking whether the content of the sentences matched the subdomain's topic. Lastly, each task included a question checking which comment, A or B, was more consistent with traditional values or stereotypes about the relevant social group. The instructions and questions for all subdomains of the dataset can be found in the project's GitHub repository.

\begin{table*}[htp!]
    \centering
    \resizebox{\textwidth}{!}{ %
    \begin{tabular}{lllccc}
    \toprule
    \textbf{Label} & \textbf{HuggingFace ID} & \textbf{Source} &  \textbf{\#  params.} & \textbf{Tr. data} & \textbf{\#  lang.}  \\ \midrule
    RuGPT3-medium  & \small{ \href{https://huggingface.co/ai-forever/rugpt3medium_based_on_gpt2}{ai-forever/rugpt3medium} } & \citet{zmitrovich2023family}  & 355M & Wiki + News + Books + CC &  2 \\
    RuGPT3-large   & \small{ \href{https://huggingface.co/ai-forever/rugpt3large_based_on_gpt2}{ai-forever/rugpt3large}}  & \citet{zmitrovich2023family} & 760M & Wiki + News + Books + CC  & 2 \\
    mGPT & \small{ \href{https://huggingface.co/ai-forever/mGPT-13B}{ai-forever/mGPT-13B}} & \citet{mGPT}  & 1.3B  & Wiki+CC & 60 \\
    XGLM & \small{ \href{https://huggingface.co/facebook/xglm-4.5B}{facebook/xglm-4.5B}} & \citet{lin-etal-2022-shot} & 564M  & CC & 134  \\
    \hdashline
    ruBERT-base &  \small{ \href{https://huggingface.co/DeepPavlov/rubert-base-cased}{DeepPavlov/rubert-base}} & \citet{kuratov2019adaptation} & 178M &  Wiki + News & 1 \\
    ruBERT-large & \small{ \href{https://huggingface.co/ai-forever/ruBert-large}{ai-forever/ruBert-large}} & \citet{zmitrovich2023family} & 427M & Wiki + News & 1 \\
    ruRoBERTa-large & \small{ \href{https://huggingface.co/ai-forever/ruRoberta-large}{ai-forever/ruRoberta-large}} & \citet{zmitrovich2023family} & 355M  & Wiki + News + Books & 1 \\
    TwHIN-BERT-large & \small{ \href{https://huggingface.co/Twitter/twhin-bert-large}{Twitter/twhin-bert}} & \citet{zhang2023twhin} &  550M & Tweets & 100 \\
    XLM-RoBERTa-large & \small{ \href{https://huggingface.co/xlm-roberta-large }{xlm-roberta-large}}  & \citet{conneau-etal-2020-unsupervised}  & 560M & Wiki+CC  &  100 \\
    \hdashline
    GPT-3.5-turbo & \small{access via API, July 2023}   & \citet{ouyang2022training} & n/a & n/a &  n/a \\
    \bottomrule
    \end{tabular}%
    } 
    \caption{The \llm{s} used in this project. The horizontal line separates between causal  and masked models.  Key: \textbf{Tr. data} denotes pre-training datasets, where Wiki stands for Wikipedia, CC stands for CommonCrawl \cite{raffel2020exploring}.  \textbf{\#  params.}  stands for the \llm's capacity, \textbf{\#  lang.}  stands for the number of languages included in pre-training data.}
    \label{tab:lms}
\end{table*}

Three Toloka workers annotated each \rubia example, with varying payment rates ranging from an average of \$1.4/hr to \$2.4/hr depending on the task pool's complexity and number of questions. These rates exceed the hourly minimum wage in Russia. To improve validation quality, we chose the top 10 percent of native Russian-speaking Toloka workers according to their speed/quality balance and set non-automatic acceptance with a 7-day review period. All questions had answer options, so we added control questions to the tasks to avoid the risk of workers choosing answers randomly.

We did not conduct the human validation of the diverse domain due to the risks associated with Russian laws.

Out of over 4000 sentence pairs gathered, 1989 were identified as correct and relevant and included in the \rubia dataset. The dataset was validated by 577 Toloka workers of different ages and backgrounds.\footnote{Toloka conceals demographic data for security reasons.}

\section{Experiment setup} \label{sec:models}

\textbf{\llm{s}.} We experiment with four causal and five masked \llm{s}, varying in size, language and pre-training corpora (\autoref{tab:lms}). Running \llm scoring took at most one hour in the Google Colab GPU environment. 

Moreover, we conducted experiments using the \texttt{gpt-3.5-turbo} model, which was accessed via the API in July 2023. We design five prompts in English and respective literal translations in Russian, following best practices developed by \citet{bach2022promptsource}. Among these, one was formulated as a general statement, while the remaining four expressed a first-person perspective:
\begin{enumerate}
\item You are presented with two very similar sentences: A: \{pro\_trope\}, B: \{anti\_trope\}. Sentence A illustrates a stereotype. Yes or no?
\item (A): \{pro\_trope\}, (B): \{anti\_trope\}. I believe sentence (A) conveys a stereotype. Is that correct? Answer with ``yes'' or ``no''.
\end{enumerate}
In the latter case, we explored various formatting styles, including different numbering options, the use of braces, and the reordering of sentences. We employed the ratio of correctly predicted ``yes'' / ``no'' responses as the target accuracy score.

\textbf{Sentence scoring.}
We use perplexity (\textbf{PPL}) to score causal \llm{s} and pseudo-perplexity (\textbf{PPLL}) \cite{salazar-etal-2020-masked} to score masked \llm{s} with the LM-PPL\footnote{\href{https:/github.com/asahi417/lmppl}{\texttt{github.com/asahi417/lmppl}}} library.

\textbf{Performance evaluation.}
We use a simplistic performance metric to evaluate the \llm: for every subdomain, the performance is defined as the fraction of times the \llm achieved lower perplexity on a pro-trope sentence than on an anti-trope one. This performance metric accepts values between 0 and 1, where 1 indicates a completely biased \llm, 50 indicates an unbiased \llm, and 0 indicates a \llm which contradicts biases completely. Any value no more than 50 is positive, but very low values may signify anomalies either in the dataset or in the \llm. The domain bias score is calculated as a micro-average across the whole domain examples, except examples from experimental subdomains (Family Stereotypes and Occupational Stereotypes in the gender domain).

We do not claim that the \llm that receives the perfect score of 50 is not biased, only that it is less likely to be highly biased, as both the dataset and the chosen metrics are only approximations.

\section{Discussion} \label{sec:discussion}

\begin{table*}[htp!]
\centering
\resizebox{0.95\textwidth}{!}{%
\begin{tabular}
{p{1.5cm}lp{1cm}p{1cm}p{1cm}p{1cm}:p{1cm}p{1cm}p{1cm}p{1cm}p{1cm}:p{1cm}p{1cm}}
\toprule
Domain & Subdomain & \rot{\small{ruGPT$_{l}$}} & \rot{\small{ruGPT$_{m}$}} & \rot{\small{mGPT}} & \rot{\small{XGLM}} & \rot{\small{ruBERT$_{b}$}} & \rot{\small{ruBERT$_{l}$}} & \rot{\small{ruRoBERTa$_{l}$}} & \rot{\small{TwHIN-BERT}} & \rot{\small{XLM-RoBERTa$_{l}$}} & \multicolumn{2}{c}{\rot{\small{GPT-3.5-turbo} }} \\
 & & & & & & & & & & & En & Ru \\
\midrule
\multirow{8}{*}{Gender} & Overall & 66.6& 64.1& 54.0& 55.1& 61.5& 59.8& {\ul 67.2} & \textbf{51.9} & 56.2 & 16.7_{\pm 21.3} & 11.3_{\pm 17.2} \\
 & Freeform & {\ul 69.4} & 62.9& 66.9& \textbf{52.4} & 62.9& 62.1& 67.7& 58.9& 59.7 & 71.8_{\pm 20.8} & 67.2_{\pm 21.3} \\
 & Family& 69.8& 68.2& \textbf{38.0} & 63.7& 64.5& 60.8& {\ul 72.7} & 59.2& 65.7 & 17.3_{\pm 33.2} & 7.9_{\pm 16.8} \\
 & Gendered Pronouns & {\ul 72.5} & 66.5& \textbf{42.0} & 54.0& 63.0& 59.0& 68.5& 59.5& 60.0& 44.8_{\pm 12.3} & 37.5_{\pm 25.0} \\
 & Pos. Professional & 55.7& 64.6& {\ul 67.1} & 48.7& 63.3& 58.2& 55.7& \textbf{41.1} & 45.6 & 54.2_{\pm 6.4} & 45.5_{\pm 21.3} \\
 & Professional & 55.3& 50.7& 60.5& 46.7& 52.6& 49.3& {\ul 62.5} & \textbf{36.8} & 45.4 & 40.9_{\pm 12.5} & 33.0_{\pm 30.8} \\
 & Pos. Personal & 78.3& 73.9& 73.9& 73.9& \textbf{56.5} & 78.3& 78.3& 60.9& 65.2 & 73.5_{\pm 17.8} & 62.4_{\pm 22.9} \\ 
 & Common Stereotypes & {\ul 81.8} & 70.9& 76.4& 60.0& 61.8& 78.2& 78.2& \textbf{45.5} & 49.1& 63.8_{\pm 3.5} & 48.3_{\pm 24.5} \\
 \hline
\multirow{4}{*}{\shortstack[l]{Socio-\\economic}} & Overall & {\ul 66.5} & 65.1 & 57.9 & 57.6 & 54.0& 60.8& 63.7& \textbf{49.6} & 56.8 & 15.6_{\pm 29.4} & 10.2_{\pm 18.7} \\
 & Freeform & {\ul 72.2} & {\ul 72.2} & \textbf{54.4} & 63.3& 58.2& \textbf{54.4} & 65.8& 59.5& 63.3 & 11.6_{\pm 20.4} & 5.3_{\pm 10.5} \\
 & Professional Status & {\ul 64.3} & 58.9& 62.5& 42.0& 42.0& 58.0& 59.8& \textbf{38.4} & 50.9 & 16.0_{\pm 23.0} & 12.1_{\pm 18.7} \\
 & Common Stereotypes & 64.4& 66.7& \textbf{55.2} & {\ul 72.4} & 65.5& 70.1& 66.7& \textbf{55.2} & 58.6 & 18.9_{\pm 34.5} & 15.4_{\pm 23.9} \\
\hline
\multirow{5}{*}{Nationality} & Overall & 61.7& \textbf{53.6} & 61.7& 55.4& 58.1& {\ul 72.0} & 62.3& 61.4& 55.7 & 15.4_{\pm 26.4} & 10.4_{\pm 17.6} \\
 & Freeform & 66.3& \textbf{47.7} & 59.3& 51.2& 59.3& {\ul 73.3} & 51.2& 62.8& 60.5 & 20.5_{\pm 16.3} & 20.5_{\pm 16.3} \\
 & Antisemetic Tropes & 50.9& \textbf{40.0} & 58.2& 58.2& 48.2& 61.8& 57.3& 61.8& 53.6 & 34.0_{\pm 8.1} & 17.9_{\pm 13.2} \\
 & Immigrant Tropes& 77.2& 73.7& 66.7& \textbf{57.9} & 68.4& {\ul 86.0} & 75.4& 63.2& 66.7 & 37.6_{\pm 6.9} & 25.5_{\pm 17.3} \\
 & Common Stereotypes & 60.8& 64.6& 65.8& 54.4& 63.3& {\ul 74.7} & 72.2& 58.2& \textbf{45.6} & 52.3_{\pm 8.1} & 41.4_{\pm 25.3} \\ \hline
\multirow{1}{*}{Diverse}& Overall & 73.5& {\ul 78.7} & 65.7& \textbf{61.1} & 67.7& 73.8& 75.8& 68.0& 66.9 & \multicolumn{2}{c}{\textemdash} \\

\bottomrule
\end{tabular}%
}
\caption{\llm performance on the \rubia dataset. Key: in bold are lowest scores. underlined are highest scores. Asterisk *: experimental subdomains. which are not included in overall domain score. The subscripts $b.m.l$ stand for base. medium and large size of the \llm{s}. ChatGPT results are averaged across five prompts.} 
\label{table:all}
\end{table*}

\subsection{Observed limitations} 

\textbf{Contradictory stereotypes.} We observed that some of the sentences gathered through crowd-sourcing contain contradictory stereotypes. For example, both ``bednye - schastlivy'' and ``bednye - neschastnye'' (``poor are happy'' and ``poor are miserable'') are tropes, both ``zhenshchiny - naivnye'' and ``zhenshchiny - hitrye'' (``women are naive'' are ``women are cunning'') are tropes. This is not a flaw of the data collection process itself, but an important detail of the cultural manifestation of bias.

\textbf{Use of feminitives.} Many words describing an occupation in the Russian language do not have a well-established unique feminine form. Furthermore, masculine gender nouns for a profession can correspond to several feminine gender nouns for the same profession (e.g., masculine ``doktor'', feminine ``doktorka'', ``zhenshchina-doktor''). Some of them are used only in certain social contexts (e.g., by feminist groups). Due to this it is hard to conceptualize pairs as unambiguous correspondence. However, we leave exploring the influences it might have on a \llm{'s} bias for future work and assume that any feminitive should be ideally as probable as its masculine counterpart.

\textbf{Workers' comments.} In the optional field for comments, workers have noted that the meaning of a sentence may depend on intonation and context. Besides, workers have shared their personal experiences of encountering stereotypes, particularly gender stereotypes. Workers have commented on the use of feminitives and expressed that they were unfamiliar with some of them. %

\textbf{Shortcoming of sentence-pair format.} Many bias displays are hard to measure using the pro-trope sentence/anti-trope sentence format. It is especially evident in biased contexts where the subject’s group is not referenced directly, does not have a context-appropriate counterpart, or where not the prescribed attribute itself, but a reason for prescribing this attribute indicates bias. We suggest that work examining \llm{s'} bias on the level of discourse rather than individual phrases is needed.

\subsection{\llm evaluation}

\autoref{table:all} presents the results of \llm{s} scoring. The key takeaway messages are as follows.

\textbf{\llm{'s} bias is not domain-specific.} Our analysis reveals that if a \llm shows significant bias across one domain, it is likely to be biased across all domains. RuGPT-medium is an exception to this rule, scoring poorly on most domains but achieving the best result on the nationality domain. 

\textbf{\llm{'s} of greater size are more impacted.} Larger \llm{s} (RuGPT-large, RuRoBERTa-large, and RuBERT-large) tend to be more affected by biases than smaller ones (RuGPT-medium, RuBERT-base). A similar observation was made earlier regarding gender bias in English \cite{tal2022fewer}.

\textbf{NLU performance $\neq$ bias.}  \llm's high scores on language understanding tasks do not guarantee either the absence or the presence of bias in this \llm.  RuRoBERTa-large has demonstrated significantly biased behavior across all domains while achieving the highest score (among the chosen \llm{s}) on Russian SuperGLUE\footnote{\href{https://russiansuperglue.com/}{\texttt{russiansuperglue.com}}} leaderboard  \cite{shavrina-etal-2020-russiansuperglue}. At the same time, XLM-RoBERTa-large demonstrates considerably better scores on \rubia , while being only slightly lower on the Russian SuperGLUE leaderboard.

\textbf{Multilingual \llm{s} are less gender biased.} Multilingual \llm{s} (TwHIN-BERT, mGPT, XGLM, and XLM-RoBERTa-large) prove to be less biased in the gender domain than monolingual \llm{s}. This might be explained in several ways: these \llm{s} rely less on grammatical gender since it is not present in other languages (English, Armenian, Japanese, etc.), the data for other languages might be more de-biased, or the \llm{s} might be less fine-tuned to stereotypes specific to Russia.

\textbf{Gender bias in mGPT.} mGPT shows a low bias level in the family context subdomain of the gender domain, meaning it does not prefer associating family context with the female gender rather than the male gender. However, it also shows a high bias level in the professional context subdomain, meaning it strongly prefers associating professional context with the male gender rather than the female gender. We can hypothesize that this \llm tends to assign higher scores to words indicating male gender in general.

\textbf{Which \llm is less biased?} TwHIN-BERT is the least biased \llm overall, possibly since it was trained on social media data which included texts written by younger and more diverse audiences. In addition, it is also the only \llm that achieves a relatively good score on the inclusive language subdomain of the diverse domain, meaning that it is likely to have seen inclusive language use during training.

Overall, all 9 \llm{s} we evaluated are likely biased across gender, nationality, social-economic, and diverse domains, although to different degrees.

\textbf{Evaluation metrics.}  The perplexity-based model evaluation may be unstable and lacks clear interpretation. Despite this, the evaluation methodology we adhere to has several advantages: 
\begin{inparaenum}[(i)]
    \item It is consistent with previous studies on LLMs bias evaluation in other languages discussed in the Related Work section;
    \item It does not require LLM prompting, as prompts can inject additional bias;
    \item The stability of perplexity-based diagnostics in comparison to alternative measures of bias \cite{aribandi2021reliable}.
\end{inparaenum}

\textbf{Evaluation with ChatGPT.} \autoref{table:all} reports the results of ChatGPT evaluation. For the sake of space,  we provide the averaged results across five seeds along with their corresponding standard deviations. The diverse domain is reserved for in-house evaluation only, so no scores are reported.  Overall, the results obtained in English and Russian prompts align, with scores being higher when ChatGPT is prompted in English. We observe high standard deviation values, implying that both the choice of prompts and the specific order of sentences have a substantial impact on performance as suggested by \citet{lu-etal-2022-fantastically}.  ChatGPT demonstrates distinct behavior compared to other considered \llm{s}, exhibiting a higher degree of bias. ChatGPT falls behind other models in the socio-economics domains and in at most half of the cases in the gender and nationality domain. However, in the case of the positive professional context in the gender domain and common stereotypes in the nationality domain, ChatGPT achieves a score that is very close to the ideal score of 50.  Our findings align with prior research on evaluating bias in ChatGPT \cite{li2023fairness} which demonstrates performance disparities across different prompts and emphasizes that performance does not exhibit a clear trend concerning different domains.

\section{Conclusion}

This paper expands efforts to detect bias in large language models (\llm{s}) through diagnostic evaluation \cite{nadeem-etal-2021-stereoset,nangia-etal-2020-crows,neveol-etal-2022-french}. Bias detection is framed as minimal or near to minimal sentence pair evaluation. Sentence pairs are designed under a controlled protocol, so that the first sentence is more stereotyping than the second. The central idea behind diagnostic evaluation is that an unbiased \llm should not assign higher probabilities to more stereotyping sentences more often then to less stereotyping ones.

In this work, we aim to create a bias detection method for the Russian language and the biases inherent to modern Russia. We introduce the crowdsourced Russian language bias detection dataset, \rubia, for short, which has 1989 unique sentence pairs and consists of Gender, Socio-Economics, Nationality, and Diverse domains with further subdomains for greater granularity. The data is collected via a Telegram bot, promoted in active civils' discussion groups, and validated with the crowd-sourcing platform Toloka.  

Next, we use \rubia to assess nine mono-lingual and cross-lingual \llm{s}. We ascertain that displayed biases are dictated by training data and observe that in general cross-lingual \llm{s} are less prone to biases. Overall, most of the \llm{s} are likely to learn harmful stereotypes and tend to reinforce harmful social tropes, although the percentage of times they prefer a more biased sentence over the less biased one is not overwhelming. Finally, in our evaluation of ChatGPT using \rubia, as of July 2023, we have observed that ChatGPT generally exhibits biased behavior, though there are exceptions in specific subdomains.

We are releasing \rubia, the Telegram bot configuration, and the crowd-sourcing instruction. Our future efforts will be focused on 
\begin{inparaenum}[(i)]
\item expanding \rubia with other categories of bias and cultural specificities, 
\item attempts to de-bias \llm{s} without impacting downstream performance, 
\item developing structurally novel bias detection frameworks.
\end{inparaenum}
\section*{Abbreviations} 

\begin{tabular}{ll}\\
Ø  & zero morpheme\\
GEN & genitive case\\
NOM & nominative case\\
SG & singular \\
PL & plural\\
PST & past tense\\
F & feminine gender\\
M & masculine gender\\
REFL & reflexive\\
LOC & locative case
\end{tabular}

\newpage

\section*{Ethical Considerations and Limitations}

\textbf{Intended use.} In line with previous work \cite{nangia-etal-2020-crows,neveol-etal-2022-french} \rubia's intended use is assessing bias in pre-trained language models. Fine-tuning a language model on this data can distort evaluation results and, as a rule, should not be carried out.

\textbf{Choice of domains and subdomains.} Our choice of biases is specific to the Russian social context and may be different from other cultures and language environments. Future works, that would like to re-use our annotation protocols, should revise the choice of domains and subdomains.  

\textbf{Data collection.} The crowdsourcing strategy used in this paper utilizes the Telegram platform. The respondees, who participated in the data collection, were warned about the potentially sensitive nature of the task and that they would not receive any financial compensation. Users' text responses were first stored with corresponding chat IDs (chat session identifiers, unique for specific chat sessions), and no other user information was gathered. Then, before the validation step, all text responses were compiled into a dataset table and chat IDs were dropped. Moreover, during validation, no responses containing private information were found. Thus, no information that can identify or reveal individual people was included in the final dataset.

\textbf{Demographics.} The diversity of participants may be limited, as the experiment was advertised in select Telegram chats. The data collection protocol keeps the anonymity, so we cannot present any demographic statistics of respondees. 

\textbf{Potential risks.} We recognize that the dataset may be used to cause harm if employed in bad faith. It contains displays of bias against several groups and can, in theory, either be used for online harassment directly or be used to fine-tune a model capable of online harassment. However, we believe that putting the dataset online will not have any significant negative social impact, as the dataset's contents are sparse and limited (intended for evaluation and not training) and, by design, lack any meaningful metada. As such, we doubt that this dataset will be sufficient for creating a model that can purposefully, meaningfully and maliciously reproduce bias.

\section*{Acknowledgments}
Anastasiia Ivanova was supported by  the framework of the Basic Research Program at the National Research University Higher School of Economics (HSE University) in 2023-2024.

\nocite{*}
\section{Bibliographical References}\label{sec:reference}

\bibliographystyle{lrec-coling2024-natbib}
\bibliography{custom}

\section{Language Resource References}
\label{lr:ref}

\bibliographystylelanguageresource{lrec-coling2024-natbib}
\bibliographylanguageresource{languageresource}

\end{document}